# Revisiting IM2GPS in the Deep Learning Era

Nam Vo
Georgia Tech
namvo@gatech.edu

Nathan Jacobs
University of Kentucky
jacobs@cs.uky.edu

James Hays
Georgia Tech
hays@gatech.edu

## Abstract

*Image geolocalization, inferring the geographic location of an image, is a challenging computer vision problem with many potential applications. The recent state-of-the-art approach to this problem is a deep image classification approach in which the world is spatially divided into cells and a deep network is trained to predict the correct cell for a given image. We propose to combine this approach with the original Im2GPS approach in which a query image is matched against a database of geotagged images and the location is inferred from the retrieved set. We estimate the geographic location of a query image by applying kernel density estimation to the locations of its nearest neighbors in the reference database. Interestingly, we find that the best features for our retrieval task are derived from networks trained with classification loss even though we do not use a classification approach at test time. Training with classification loss outperforms several deep feature learning methods (e.g. Siamese networks with contrastive of triplet loss) more typical for retrieval applications. Our simple approach achieves state-of-the-art geolocalization accuracy while also requiring significantly less training data.*

## 1. Introduction

In recent years, the recognition community has broadened its focus beyond object categorization to the understanding of a litany of object, scene, material, or 3D attributes. One of the most important attributes of an image is *geolocation* – if we know the location of a photo, we trivially know hundreds of additional attributes (any attribute for which a map exists, e.g. population density, average temperature, crime rate, elevation, distance to a McDonald's, etc.). Knowing the location of an image is also a common photo forensics task. For example, the mobile app *TraffickCam* collects hotel room images to locate incidents of abuse. Unlike many computer vision tasks, computational systems typically exceed the performance of humans at image geolocalization because it is hard for humans to have an accurate visual model of the entire world.

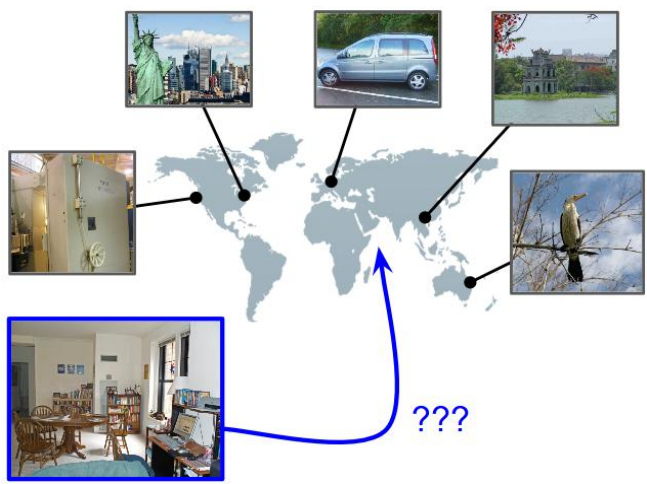

Figure 1. This work addresses the image geolocalization problem: given a large set of GPS-tagged images, learn to infer the GPS coordinate of a query image with unknown location.

Estimating the geolocation of an arbitrary photo is still a challenging task (Figure 1). In particular, we examine the task of predicting the location of a single photo given only the image content with no metadata. We consider this task at a global scale and attempt to estimate the GPS coordinates for any query image. For this task, localization can be considered successful if the estimated location is within a specified error threshold. Depending on the application, this threshold could be street level (1km), city level (25km), region level (200km), country level (750km), or continent level (2500km). We adopt these five levels of granularity from prior work and examine the performance of geolocalization strategies at these error thresholds.

One natural approach to the image geolocalization task would be to to treat it like an *instance retrieval* task and match local features from the query image (and perhaps their geometric layout) to a reference database of images with known locations [16]. Such approaches work well if (1) there are images in the reference database with a field of view that significantly overlaps with that of the query image and (2) if the content of the query image is well suited to local feature matching (i.e., it has distinctive man-made or geological features). Unfortunately, this is often not the case, especially for query images away from tourist destinations and dense urban areas. Therefore, it is necessary to infer location without requiring direct local-feature matching. In these cases, image geolocalization is similar to *scene classification* or *scene attribute estimation* in that a system needs to achieve a higher-level, more qualitative understanding of an image, e.g. recognizing that buildings are typical of Greek islands even though this particular island isn't in the reference database.

Im2GPS [9, 10] introduced the global geolocalization problem and used hand-crafted features from the *instance recognition* and *scene classification* literature jointly to retrieve nearest neighbors in a database of 6 million geotagged images. Im2GPS found that roughly half of successful geolocalizations are instance level matches whereas half are more qualitative matches based on shared scene attributes (geology, architecture, land cover, etc.).

More recently, PlaNet [36] formulates image geolocalization as a classification task. This is done by mapping the GPS coordinate (a pair of real numbers) to a discrete class label by dividing the surface of the earth into distinct regions. PlaNet proposes a deep convolutional neural network to estimate a probability distribution over regions from raw pixel values. PlaNet not only significantly outperforms Im2GPS in terms of accuracy, it is also dramatically faster since it requires only a forward pass through a deep network instead of a nearest neighbor search through millions of image features.

Is the deep image classification formulation of PlaNet the best approach to geolocalization (as it seems to be for most other scene understanding tasks)? There are two reasons to suspect it might not be ideal – first, discretizing the Earth's surface is lossy since we are ultimately interested in a real-valued location estimate (potentially expressed through GPS coordinates). Second, and more limiting, is that a single deep network, even with tens of millions of parameters, will struggle to memorize the visual appearance of the entire Earth. An effective deep network needs to learn to do both *instance* matching and more *qualitative* scene understanding. Can contemporary deep networks implicitly 'memorize' tens of millions of photographic features necessary for the instance matching?

In this paper, we adopt the retrieval approach of Im2GPS but pair it with deep feature learning as in PlaNet. We outperform PlaNet by a significant margin – **47.7% accuracy vs 37.6% for PlaNet** on the Im2GPS test set with a 200km threshold . Interestingly, while we approach geolocalization as a retrieval task with learned deep features, we don't see a benefit to using embedding formulations (e.g. Siamese networks with contrastive or triplet loss) typical for retrieval tasks. Our best performance comes from training a classification network, in the spirit of PlaNet, and using its intermediate activations as our image feature.

The contributions of this study are:

- We significantly improve the state-of-the-art accuracy for global image geolocalization. Our increase in accuracy is similar in magnitude to that achieved by PlaNet [36] over the hand-crafted retrieval approach of Im2GPS. We achieve this with as little as 5% of the training data used by PlaNet, and increase the gap further while using 28% as much reference data.

- Our increase in accuracy comes from changing the formulation from classification to retrieval. The benefit of retrieval in this setting is a reflection of the geolocalization problem and the nature of current deep models – the visual world is too complex for a deep model to memorize, but a retrieval approach does so trivially.

- We investigate different strategies for learning a deep feature embedding for geolocalization. Surprisingly, deep feature learning methods typically used for retrieval applications do not outperform training with a classification loss. For classification-based localization, we find that different discretization strategies also have a significant impact.

- Through extensive experimentation, we find that some training procedures lead to higher accuracy at the street scale (1km) and others at the country scale (750km). We observe a trade off between fine-scale and coarse-scale performance, the regimes traditionally approached with instance-level matching methods and scene classification methods, respectively.

Related works are discussed in the next section. We describe image geolocalization system designs in Section 3. Experiments and analysis are reported in Section 4 and we conclude in Section 5.

## 2. Related Work

Recent years have seen a dramatic expansion of deep learning methods for scene understanding tasks [14]. Deep learning has been applied successfully to location prediction [36] and other tasks related to our problem: predicting scene type [40], perceptual attributes [6] such as safety, liveliness and geo-informative attributes [15] like GDP, elevation.

Image retrieval using learned, deep representations is useful to a wide range of tasks such as product ranking [4, 11], sketch based image retrieval [25], face recognition [31, 26], cross-view localization [18, 37, 34] and scene retrieval [33, 2, 23, 7]. Distance metric learning (DML) is usually employed with a deep network, most commonly using the contrastive loss [8] or triplet ranking (hinge loss) [5, 19, 35]. New loss functions and example mining strategies have been being proposed as they play important role in the learning process [24, 20, 23, 34].

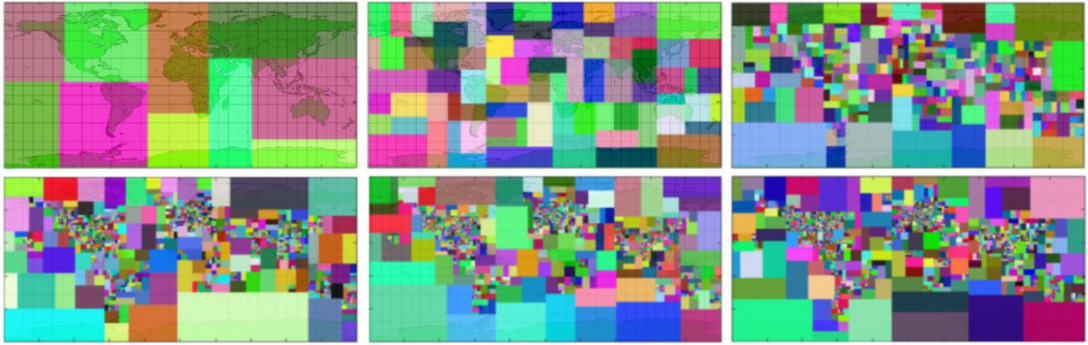

Figure 2. We study six schemes for discretizing geographic location. These vary from coarse to fine (10, 80, 359, 1060, 1693 and 7011 regions respectively).

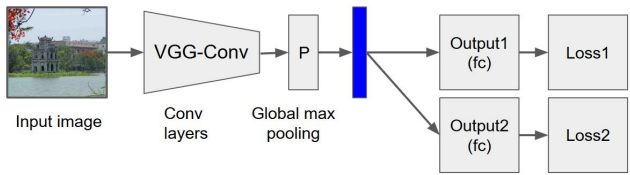

Figure 3. Our proposed CNN architecture consists of the convolutional layers of the VGG-16 network [28] followed by a global max pooling layer. Depending on the task, we append to this an output layer and the corresponding loss layer. For classification, we use a fully connected layer and Softmax-CrossEntropy loss, for retrieval, we use a DML loss.

We are studying image retrieval geolocalization which has overlap with *instance-level scene retrieval* [33, 2, 23, 7]; Since this line of work mostly focuses on instance-level matching, benchmarks designed for this task consist of popular scenes or landmarks [21, 22, 12] and similarity between matched images are visually recognizable (by humans or geometric verification). In this regime, with manual labeling and/or clever example mining, it is beneficial to apply distance metric learning. Techniques such as geometry verification or query expansion typically improve instance retrieval mAP, but these techniques are less useful when geolocalizing scenes that do not have instance matches.

Many previous works on image localization are at limited spatial scale (urban areas) or on special class of images (landmarks, streetview) [16, 39, 18, 34, 1, 17, 38]. Many approaches make used of aerial imagery for localization [3, 3, 27, 17]. In [18, 34, 37], images of the same scene from the ground viewpoint and overhead viewpoint are embedded in the same feature space through deep learning DML; the resulting system then does localization by image retrieval using reference database of aerial images.

Also related, to match aerial images across wide baselines, Altwaijry et al. [1] propose a deep attentive architecture to classify whether two views match.

Image geolocalization at planet scale is challenging and less studied – only Im2GPS [9, 10] and PlaNet [36] aim for global coverage. These are the most closely related works on we build on both.

## 3. Image Geolocalization using Deep Learning

Given a large training data of images with GPS labels, we examine two deep learning approaches for geolocalization. For both cases we use the same architecture shown in Figure 3 which has been popular for landmark recognition [2, 23, 7].

### 3.1. Geolocalization by classification

One approach is to formulate geolocalization as a classification problem [36]: the GPS label is converted to class label by quantizing all GPS labels to a fixed number of classes, so that each class represents a physical region in the real world. The classification result then can be converted back to the GPS coordinate of the corresponding region.

PlaNet[36] divides the Earth into a set of geographical cells based on image density. We derive a similar adaptive scheme: starting with a single cell of the entire world, repeatedly divide each cells along latitude or longitude whichever side is bigger (either evenly or randomly) until the number of images in each cell is smaller than a threshold $t_{img}$ or the physical area is smaller than a threshold $t_{area}$; these parameters define how fine the partitioning is.

To predict the location as precisely as possible, one would prefer a fine-grained partitioning (for example [36]'s partitioning has 26,263 cells). However we should take into account the training data's size, the learning model's capacity and especially the localization error tolerance. We investigate 6 different partitionings shown in Figure 2. Admittedly these choices are somewhat arbitrary, as we do not directly control the number of cells, nor do we try to "optimize" each partitioning. We used similar parameter to [36] to obtain a fine grained partitioning (though [36]'s data is ∼14 times bigger so they still have $\sqrt{14}$ times more cells); then we loosen the thresholds to obtain the other 5 coarser partitionings.

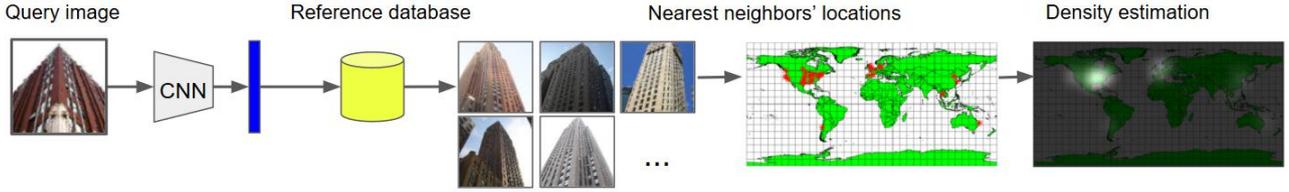

Figure 4. A visual overview of our image-retrieval approach to image geolocalization. We extract a feature from our CNN, find nearby neighbors in feature space, and estimate the GPS coordinate using either the top NN or the density.

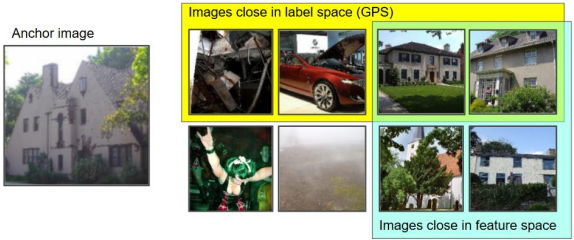

Figure 5. When performing distance metric learning, we sample images based on their distance, either in label space (geographic distance) or feature space, to an anchor image. Some example images that are close/far from an anchor image in the label space/feature space.

**Multiple class labeling**: We investigate the effect of using multiple partitionings simultaneously. The motivation is that different proximity information is preserved at different levels of granularity (and not the others). Moreover classification results at multiple coarse partitionings can be combined to produce a more fine grained prediction. Therefore we experiment with training multiple classification losses as these tasks are heavily correlated and benefit each other.

### 3.2. Geolocalization by image retrieval

This approach looks up images that are similar to the query images and makes use of the known locations of those images [9]. This requires learning a representation for comparing images (for which we will use deep learning) and indexing a large reference database.

To learn such a representation, we employ distance metric learning (ranking/triple hinge loss, contrastive loss and similar loss functions) which requires pairs of images labeled 'similar' or 'different'. When not available, such labeling can be automatically generated using geometry verification [7, 23] or class labels [20, 24]. In our case, we make use of the class label described in the previous section or directly threshold the GPS distance between the 2 images. Similar to [2], we can also match images that are not only close in the GPS label space but also close in the current feature space. Even so, with the data we are dealing with, this supervision is very weak in the sense that matched images (taken at the same location/region) are most likely not of the same or even similar scene/object (Figure 5).

After training we use the CNN as a feature extractor and index a large dataset of reference image features. At test time, we look up the nearest neighbor (NN) of the query image in the feature space using approximate NN search and output its location (Figure 4). This approach works based on the assumption that, after learning, images close in the feature space are likely to be close in the label (GPS coordinate) space too.

**k-NN density estimation**: we can make use of the top k NN instead of only 1. We perform weighted kernel density estimation using each NN as a Gaussian kernel, the density at a point $x$ in GPS coordinate space can be written as:

$$f(x) = \sum_{i=1}^{k} w_i N(x; x_i, \sigma^2 I) \quad (1)$$

Where $x_i$ is the GPS coordinate of the i-th NN, we also weight each NN $w_i = s_i^m$ depending on its similarity score $s_i$ (defined to be the inverse of the distance between the query image's feature and the reference image's feature). The point with highest density is chosen as output.

Note that as k decrease, m increase or $\sigma$ decreases, this output becomes the NN. These parameters can be optimized: bigger reference data allows bigger k and looser error threshold allows bigger $\sigma$. Given our dataset (described in the next section) We choose $m = 10$, $k = 100$ through validation (these parameters were not precisely tuned) and experimentally manipulate $\sigma$.

## 4. Experiments

**Training data**: We use the Im2GPS dataset from [9]. It consists of more than 6 million images collected from Flickr that are tagged with countries or states' name and also have GPS coordinates. This data is used for GPS quantization

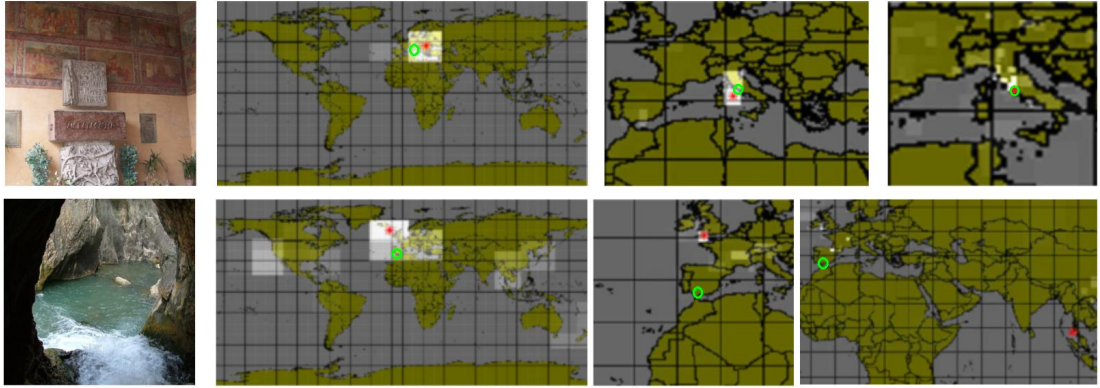

Figure 6. Example results of geolocalization by image classification using different partitionings. From left to right: input images, classification result with 80, 1060 and 7011 classes respectively (lighter region means higher probability). Red * denotes the predicted location and green o denotes the true location.

(Figure 2), training deep networks, and as retrieval reference database.

**Testing data**: for analysis, we construct 2 test sets; we make sure that no image from training and test data come from the same photographer.

- **Im2GPS3k**: 3000 images from Im2GPS. Note that this is different from the Im2GPS test set [9].
- **YFCC4k**: 4000 random images from the YFCC100m dataset [32]. Since it is designed for general computer vision purpose, its image distribution is different from Im2GPS making this test set more challenging.

**Training for classification**: we train the following networks:

- **L**one: We trained a network with a single classification loss corresponding to the most fine grained partition (7011 classes). This can be considered an analog of PlaNet [36] at smaller scale. We also train another version **L2** for the 359 ways classification loss.
- **M**ulti: We train another classification network with 6 different losses corresponding to 6 partitions scheme described in section 3.1. Hence this network produces 6 localization outputs . We'll treat these outputs independently and evaluate the performance of each of them.

**Training for retrieval**: we fine-tune the model L with ranking loss (triplet hinge loss) to learn a better representation, resulting in a **R**anking network. To do localization by retrieval, we experiment with different networks as feature extractor: the classification networks (L and M) and the ranking network (R).

We also evaluate two other publicly available state-of-the-art models, NetVLAD[2] and Siamac[23], which have similar architecture (VGG-conv layers), but different training data (weakly supervised Google streetview time machine and SfM landmark images hard example mining), global pooling layer (NetVLAD and R-Max) and loss function (triplet hinge loss and contrastive loss). Different from our approach, these models have an additional fully-connected layer for PCA. Features from all models are L2-normalized when used for retrieval.

**Notation**: we will use $[Model]Approach$ to refer to each method, where $Model$ can be L, L2, M, R, NetVLAD, Siamac described above, and $Approach$ can be C (for classification), NN, kNN (for retrieval). For example [M]311C refers to the 311 way classification output of model M, and [M]NN refer to the NN retrieval approach using model M as feature extractor.

**Metric**: the geolocalization accuracy is defined as the percentage of test images whose predicted location is within the error threshold from the true location. Similar to [9, 10, 36], 5 error thresholds are used: 1km, 5km, 25km, 750km, 2500km corresponding to 5 levels of localization: street, city, region, country, continent.

**Result**: Qualitative results are shown in Figures 6 and 7. Quantitative results on two test sets are shown in Figure 8. For comparison we add a simple baseline: always outputting **London**, which is the region with the most images. This baseline is practically the best one can do without looking at the input image; its performance is much better than guessing a random location on the Earth.

We will ensure that our results can be replicated by sharing our datasets, source code, and trained models.

### 4.1. Comparing classification performance

An example output of classification is in Figure 6. In the case of less ambiguous image, the network would be able to predict the correct region/cell. Since the center of the region is used, a finer partitioning will lead to a prediction that is closer to the true location (top row). Though in case of the image being very ambiguous, correctly localizing it at coarser level is more likely (bottom row).

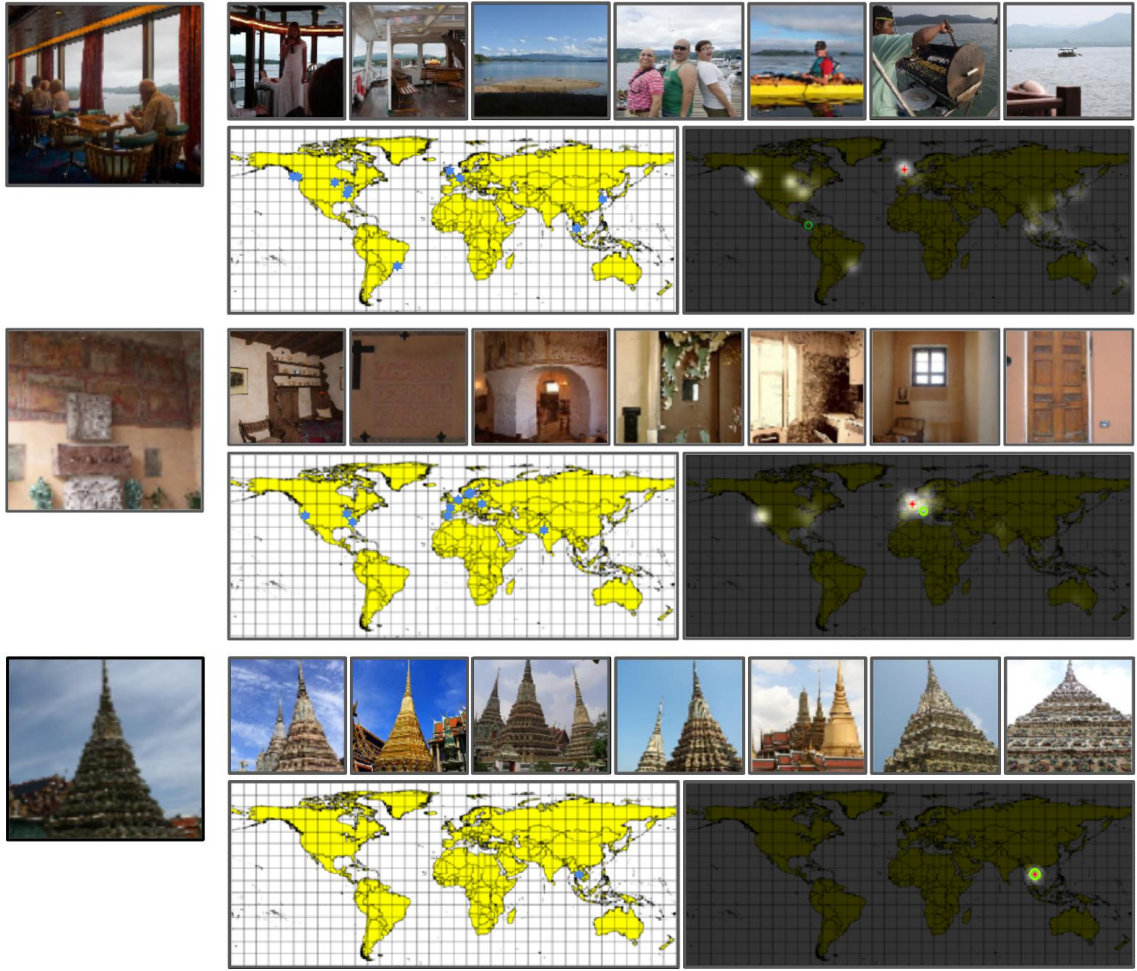

Figure 7. Example results of our geolocalization by image retrieval system (kNN, $\sigma=4$). Each row shows the input image on the left, the first few NNs on the right, together with their locations (blue *). At the end of the row we show the density result, red * denotes the predicted location and green o denotes the true location.

As shown in Figure 8, the geolocalization accuracy of the 10 way classification output is quite bad, this is mostly because at this scale the Earth is under-divided. We can see that as the partitioning is finer, the localization performance at lower error threshold gets better as expected. The fine-grained classification output (7011C) outperforms others at street and city level.

Most interesting, the geolocalization accuracy at coarse level gets worse if the partitioning is too fine: for example at continent level, the 80C and 359C achieve highest accuracy; At country level, the 359C and 1060C have the advantage. This seems to indicate a trade off between the accuracy at coarse and fine level, which may be a shortcoming of the partitioning in PlaNet [36].

[M]7011C and [L]7011C achieve similar accuracy ([L] is slightly better). However in the case of 359C, [M] is slightly better than [L2]. This suggests that when training with multiple classification losses, the fine-grained one seems to help the coarse one a little, but not vice versa.

### 4.2. Comparing retrieval performance

Figure 7 shows example image retrieval results. The NNs are similar scenes to the input image. In the case of landmarks and popular sites, they are usually instance level matches.

As shown in Figure 8, with localization by NN image retrieval, all 5 models (R, M, L, NetVLAD, Siamac) perform well and outperform the classification result at street and city level. This makes sense as these successful localizations are likely correct instance-level matches. While classification network can learn the general characteristics of each regions, it doesn't have enough capacity to 'remember' all specific instances, while the retrieval approach 'remembers' this by directly saving all reference features.

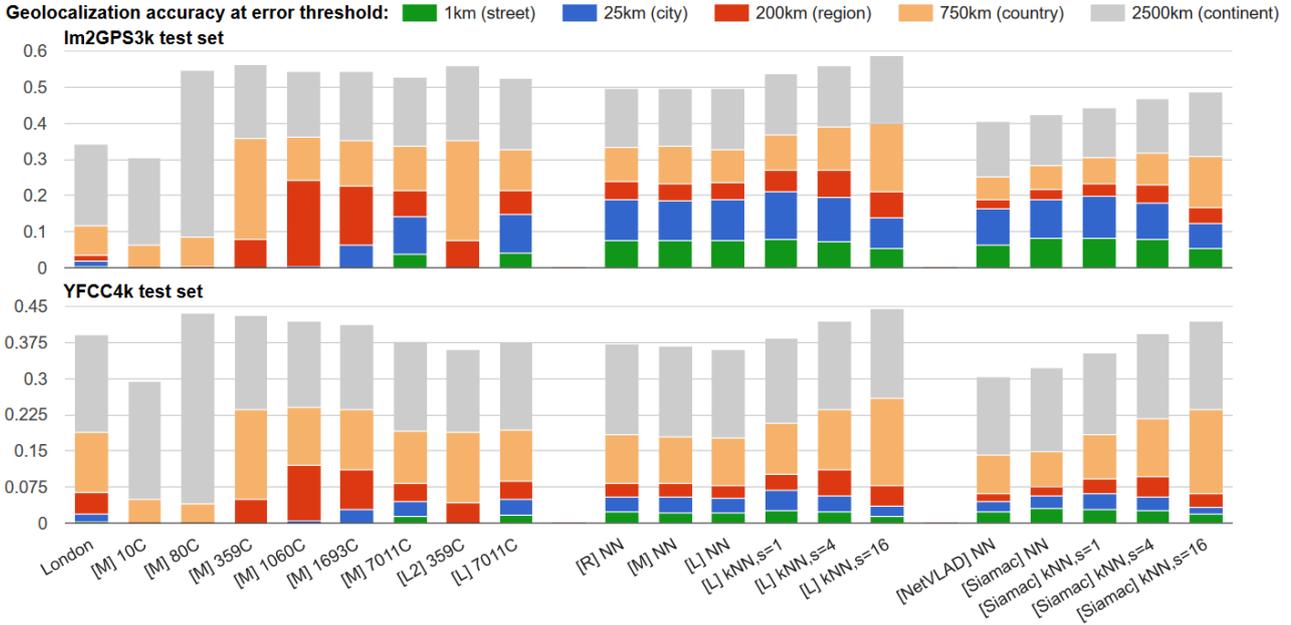

Figure 8. Geolocalization accuracy on two test sets. Note that the accuracy is presented as the top of the bars, not the length of each single color.

Among all 5 models, NetVLAD is the worst. Siamac is the most discriminative at street level. As a trade off, it has slightly lower performance at coarse level (country and continent). The L and M models are comparable and they perform relatively well even though they are trained for classification. Coarse partitioning classification approaches still have the advantage at country and continent scale.

Finally, using kNN-kernel density estimation improves the accuracy (here we only show [L] and [Siamac] but the changes when using other models are similar); especially at coarse scales (as $\sigma$ increases) this makes retrieval competitive with the classification approach. However bigger $\sigma$ can potentially lower the accuracy at fine grained level. Interestingly, we arrive at a similar trade off between fine and coarse geolocalization accuracy.

### 4.3. Training a ranking network with GPS label

Model R (which was fine-tuned from L) doesn't produce a noticeable improvement over L or M (Figure 8). In further investigations, we train a dozen versions of R, fine-tuned from different pretrained models and varied the way we sample/mine training examples. In all cases, little progress is observed in term of both training loss and geolocalization performance.

However when using landmark matches from [23] for training instead of Im2GPS data, we observe slight improvement at street level, but worse results at other scales. This is consistent with the fact that Siamac[23] is very good at street level.

Distance metric learning losses like triplet hinge loss seems to be very sensitive to noisy labels. Different from classification loss (where the label for each image is fixed during training), the "target" of each training image keep changing while they are adjusting distance from each other, usually making convergence slower.

We hypothesize that the inter-class ambiguity and intra-class diversity are too large and DML is not able to learn from GPS supervision (Figure 5).

### 4.4. Comparing with IM2GPS and PlaNet

On the Im2GPS test set, we can directly compare Im2GPS [9, 10] and PlaNet [36] with two of our models:
- The fine-grained classification network ([L] 7011C). This can be considered the equivalent of Google's PlaNet[36] at smaller scale.
- kNN kernel density estimation retrieval ([L] kNN, $\sigma$=4). This can be considered the equivalent of Im2GPS approach [9], but using deep features instead of classical features.

The result is shown in table 1. Our classification network outperforms Im2GPS even though it is still not as good as PlaNet. On the other hand, our localization by deep learnt image retrieval method produces even better accuracies. This result highlights the advantage of retrieval approach for fine-grain localization.

**Complexity analysis:** in term of number of parameters without counting the output layers, PlaNet is 3 times bigger than our 13 layers deep VGG model. Note that PlaNet uses an Inception architecture which has been heavily designed to optimize for complexity [29, 30] (for reference, it is 8 times bigger than 22 layers deep GoogLeNet[29] and 2 times bigger than 42 layers deep InceptionV2[30]). Also PlaNet's training data has more than 90 million images and it takes 2.5 months to train on clusters (approximately 40 years of CPU time). However in term of space complexity, our image retrieval approach requires all reference features be available during testing, not just the deep network. Moreover, the cost of indexing and perform NN search is not negligible; though indexing needs to be done only once and in our experience the cost of approximate NN search is smaller than that of feature extraction.

Comparing to Im2GPS [9, 10], deep learning feature extraction is orders of magnitudes faster than computing many classical computer vision features. Im2GPS's combined feature has more than 100k dimensions; in [10] lazy learning is done for each query adding more time complexity. In contrast, our deep feature with 512 dimensions is suitable for direct comparisons in Euclidean space. Because of this, our kNN kernel density estimation is a more efficient and effective post-processing procedure than the similar kNN mean shift clustering and lazy learning in [10].

### 4.5. Effect of retrieval reference database

One advantage of retrieval approach is that we can simply index more examples to improve the performance. To that end we collect another 22 million GPS-tagged images from the YFCC100m dataset [32], increasing our database size to a total of 28 million images. As shown in table 1 (last row), this results in better performance of [L]kNN,$\sigma$=4 on the Im2GPS test set.

We vary the reference retrieval database (Im2GPS-6 millions images, YFCC-22 millions and the combined 28 million) and show the geolocalization accuracy in table 2. The performance when using YFCC22m is actually no better than when simply using Im2GPS; though the combined database of 28 million images result in an improvement. We attribute this to the fact that the IM2GPS test set and the IM2GPS database come from the same distribution, which makes IM2GPS more useful for referencing. To quantify this, we measure the percentage of IM2GPS images among the top 1, 10, 100, 1000 nearest neighbors result, they are 53.2%, 50.1%, 44.6% and 40.1% respectively, which is quite high given that IM2GPS only constitutes 22.8% of the combined database.

Similar to result on Im2GPS3k and YFCC4k, we can change $\sigma$ to optimize the accuracy at a localization level (at the expense of the others). If the system is allowed to produce different outputs at different levels, this further out-performs the result in Table 1.

Table 1. Performance on Im2GPS test set. (Human* performance is average from 30 mturk workers over 940 trials, so it might not be directly comparable)

| Threshold (km) | Street 1 | City 25 | Region 200 | Country 750 | Cont. 2500 |
|---|---|---|---|---|---|
| Human* | | | 3.8 | 13.9 | 39.3 |
| Im2GPS [9] | | 12.0 | 15.0 | 23.0 | 47.0 |
| Im2GPS [10] | 02.5 | 21.9 | 32.1 | 35.4 | 51.9 |
| PlaNet [36] | 08.4 | 24.5 | 37.6 | 53.6 | **71.3** |
| [L] 7011C | 06.8 | 21.9 | 34.6 | 49.4 | 63.7 |
| [L] kNN, $\sigma$=4 | **12.2** | **33.3** | **44.3** | **57.4** | **71.3** |
| ... 28m database | 14.4 | 33.3 | 47.7 | 61.6 | 73.4 |

Table 2. Performance on Im2GPS test set based on different retrieval reference database.

| Retrieval | Database | Stre. | City | Reg. | Cou. | Cont. |
|---|---|---|---|---|---|---|
| [L] NN | Im2GPS | 12.7 | 33.3 | 40.9 | 53.2 | 71.7 |
| | YFCC22m | 12.2 | 30.4 | 37.6 | 51.1 | 67.1 |
| | Both(28m) | 13.9 | 32.9 | 40.5 | 54.4 | 70.9 |
| [L] kNN $\sigma = 1$ | Im2GPS | 13.1 | 36.3 | 44.3 | 56.1 | 70.0 |
| | YFCC22m | 12.7 | 34.2 | 43.9 | 55.3 | 68.8 |
| | Both(28m) | **15.2** | **37.6** | 46.0 | 57.0 | 69.2 |
| [L] kNN $\sigma = 4$ | Im2GPS | 12.2 | 33.3 | 44.3 | 57.4 | 71.3 |
| | YFCC22m | 11.8 | 31.2 | 42.2 | 58.7 | 70.0 |
| | Both(28m) | 14.4 | 33.3 | **47.7** | **61.6** | 73.4 |
| [L] kNN $\sigma = 16$ | Im2GPS | 10.6 | 24.9 | 35.4 | 59.5 | 75.9 |
| | YFCC22m | 8.4 | 19.8 | 34.6 | 58.2 | 74.7 |
| | Both(28m) | 11.8 | 24.9 | 36.7 | 60.8 | **77.2** |

## 5. Conclusion

We presented a deep learning study on image geolocalization, where we experimented with several settings of image classification and image retrieval approaches adapted to this task. We do not claim technical novelty for any components of this study. Our approaches are relatively simple yet achieve state-of-the-art accuracy. In the end, the best performing models can efficiently and accurately localize at coarse level using classification, and if needed can search for instance matches using retrieval techniques.

The main goal of this paper is to investigate the effectiveness of deep learning methods for geolocalization. With the newly obtained insights, we think the following lines of future work would be important: (1) we have shown the dependency between partitioning scheme and geolocalization accuracy, which begs the question: what is the best way to partition and how can the partitioning be optimized given a particular error threshold? (2) Are GPS labels too weak a supervision for traditional deep distance metric learning? There is likely an opportunity for better weakly supervised DML to improve the geolocalization.

# References

[1] H. Altwaijry, E. Trulls, J. Hays, P. Fua, and S. Belongie. Learning to match aerial images with deep attentive architectures. In *Proceedings of the IEEE Conference on Computer Vision and Pattern Recognition*, 2016. 3

[2] R. Arandjelovic, P. Gronat, A. Torii, T. Pajdla, and J. Sivic. Netvlad: Cnn architecture for weakly supervised place recognition. In *Proceedings of the IEEE Conference on Computer Vision and Pattern Recognition*, pages 5297–5307, 2016. 2, 3, 4, 5

[3] M. Bansal, K. Daniilidis, and H. Sawhney. Ultra-wide baseline facade matching for geo-localization. In *Computer Vision–ECCV 2012. Workshops and Demonstrations*, pages 175–186. Springer, 2012. 3

[4] S. Bell and K. Bala. Learning visual similarity for product design with convolutional neural networks. *ACM Transactions on Graphics (TOG)*, 34(4):98, 2015. 2

[5] G. Chechik, V. Sharma, U. Shalit, and S. Bengio. Large scale online learning of image similarity through ranking. *The Journal of Machine Learning Research*, 11:1109–1135, 2010. 3

[6] A. Dubey, N. Naik, D. Parikh, R. Raskar, and C. A. Hidalgo. Deep learning the city: Quantifying urban perception at a global scale. In *European Conference on Computer Vision*, pages 196–212. Springer, 2016. 2

[7] A. Gordo, J. Almazán, J. Revaud, and D. Larlus. Deep image retrieval: Learning global representations for image search. In *European Conference on Computer Vision*, pages 241–257. Springer, 2016. 2, 3, 4

[8] R. Hadsell, S. Chopra, and Y. LeCun. Dimensionality reduction by learning an invariant mapping. In *Computer vision and pattern recognition, 2006 IEEE computer society conference on*, volume 2, pages 1735–1742. IEEE, 2006. 2

[9] J. Hays, A. Efros, et al. Im2gps: estimating geographic information from a single image. In *Computer Vision and Pattern Recognition, 2008. CVPR 2008. IEEE Conference on*, pages 1–8. IEEE, 2008. 2, 3, 4, 5, 7, 8

[10] J. Hays and A. A. Efros. Large-scale image geolocalization. In *Multimodal Location Estimation of Videos and Images*, pages 41–62. Springer, 2015. 2, 3, 5, 7, 8

[11] J. Huang, R. S. Feris, Q. Chen, and S. Yan. Cross-domain image retrieval with a dual attribute-aware ranking network. In *Proceedings of the IEEE International Conference on Computer Vision*, pages 1062–1070, 2015. 2

[12] H. Jegou, M. Douze, and C. Schmid. Hamming embedding and weak geometric consistency for large scale image search. In *European conference on computer vision*, pages 304–317. Springer, 2008. 3

[13] Y. Jia, E. Shelhamer, J. Donahue, S. Karayev, J. Long, R. Girshick, S. Guadarrama, and T. Darrell. Caffe: Convolutional architecture for fast feature embedding. *arXiv preprint arXiv:1408.5093*, 2014. 1

[14] A. Krizhevsky, I. Sutskever, and G. E. Hinton. Imagenet classification with deep convolutional neural networks. In *Advances in neural information processing systems*, pages 1097–1105, 2012. 2

[15] S. Lee, H. Zhang, and D. J. Crandall. Predicting geo-informative attributes in large-scale image collections using convolutional neural networks. In *Applications of Computer Vision (WACV), 2015 IEEE Winter Conference on*, pages 550–557. IEEE, 2015. 2

[16] Y. Li, D. J. Crandall, and D. P. Huttenlocher. Landmark classification in large-scale image collections. In *Computer vision, 2009 IEEE 12th international conference on*, pages 1957–1964. IEEE, 2009. 1, 3

[17] T.-Y. Lin, S. Belongie, and J. Hays. Cross-view image geolocalization. In *Computer Vision and Pattern Recognition (CVPR), 2013 IEEE Conference on*, pages 891–898. IEEE, 2013. 3

[18] T.-Y. Lin, Y. Cui, S. Belongie, and J. Hays. Learning deep representations for ground-to-aerial geolocalization. In *Proceedings of the IEEE Conference on Computer Vision and Pattern Recognition*, pages 5007–5015, 2015. 2, 3

[19] M. Norouzi, D. J. Fleet, and R. R. Salakhutdinov. Hamming distance metric learning. In *Advances in neural information processing systems*, pages 1061–1069, 2012. 3

[20] H. Oh Song, Y. Xiang, S. Jegelka, and S. Savarese. Deep metric learning via lifted structured feature embedding. In *The IEEE Conference on Computer Vision and Pattern Recognition (CVPR)*, June 2016. 3, 4

[21] J. Philbin, O. Chum, M. Isard, J. Sivic, and A. Zisserman. Object retrieval with large vocabularies and fast spatial matching. In *Computer Vision and Pattern Recognition, 2007. CVPR'07. IEEE Conference on*, pages 1–8. IEEE, 2007. 3

[22] J. Philbin, O. Chum, M. Isard, J. Sivic, and A. Zisserman. Lost in quantization: Improving particular object retrieval in large scale image databases. In *Computer Vision and Pattern Recognition, 2008. CVPR 2008. IEEE Conference on*, pages 1–8. IEEE, 2008. 3

[23] F. Radenović, G. Tolias, and O. Chum. Cnn image retrieval learns from bow: Unsupervised fine-tuning with hard examples. In *European Conference on Computer Vision*, pages 3–20. Springer, 2016. 2, 3, 4, 5, 7

[24] O. Rippel, M. Paluri, P. Dollar, and L. Bourdev. Metric learning with adaptive density discrimination. *arXiv preprint arXiv:1511.05939*, 2015. 3, 4

[25] P. Sangkloy, N. Burnell, C. Ham, and J. Hays. The sketchy database: learning to retrieve badly drawn bunnies. *ACM Transactions on Graphics (TOG)*, 35(4):119, 2016. 2

[26] F. Schroff, D. Kalenichenko, and J. Philbin. Facenet: A unified embedding for face recognition and clustering. In *Proceedings of the IEEE Conference on Computer Vision and Pattern Recognition*, pages 815–823, 2015. 2

[27] Q. Shan, C. Wu, B. Curless, Y. Furukawa, C. Hernandez, and S. M. Seitz. Accurate geo-registration by ground-to-aerial image matching. In *3D Vision (3DV), 2014 2nd International Conference on*, volume 1, pages 525–532. IEEE, 2014. 3

[28] K. Simonyan and A. Zisserman. Very deep convolutional networks for large-scale image recognition. *arXiv preprint arXiv:1409.1556*, 2014. 3, 1

[29] C. Szegedy, W. Liu, Y. Jia, P. Sermanet, S. Reed, D. Anguelov, D. Erhan, V. Vanhoucke, and A. Rabinovich.

Going deeper with convolutions. In *Computer Vision and Pattern Recognition (CVPR)*, 2015. 8

[30] C. Szegedy, V. Vanhoucke, S. Ioffe, J. Shlens, and Z. Wojna. Rethinking the inception architecture for computer vision. In *Proceedings of the IEEE Conference on Computer Vision and Pattern Recognition*, pages 2818–2826, 2016. 8

[31] Y. Taigman, M. Yang, M. Ranzato, and L. Wolf. Deepface: Closing the gap to human-level performance in face verification. In *Computer Vision and Pattern Recognition (CVPR), 2014 IEEE Conference on*, pages 1701–1708. IEEE, 2014. 2

[32] B. Thomee, D. A. Shamma, G. Friedland, B. Elizalde, K. Ni, D. Poland, D. Borth, and L.-J. Li. Yfcc100m: The new data in multimedia research. *Communications of the ACM*, 59(2):64–73, 2016. 5, 8

[33] G. Tolias, R. Sicre, and H. Jégou. Particular object retrieval with integral max-pooling of cnn activations. *arXiv preprint arXiv:1511.05879*, 2015. 2, 3

[34] N. N. Vo and J. Hays. Localizing and orienting street views using overhead imagery. In *European Conference on Computer Vision*, pages 494–509. Springer, 2016. 2, 3

[35] J. Wang, Y. Song, T. Leung, C. Rosenberg, J. Wang, J. Philbin, B. Chen, and Y. Wu. Learning fine-grained image similarity with deep ranking. In *Computer Vision and Pattern Recognition (CVPR), 2014 IEEE Conference on*, pages 1386–1393. IEEE, 2014. 3

[36] T. Weyand, I. Kostrikov, and J. Philbin. Planet-photo geolocation with convolutional neural networks. In *European Conference on Computer Vision*, pages 37–55. Springer, 2016. 2, 3, 4, 5, 6, 7, 8

[37] S. Workman, R. Souvenir, and N. Jacobs. Wide-area image geolocalization with aerial reference imagery. In *ICCV 2015*, 2015. 2, 3

[38] A. R. Zamir, A. Hakeem, L. Van Gool, M. Shah, and R. Szeliski. Large-scale visual geo-localization. *Advances in computer vision and pattern recognition (*, 2016. 3

[39] A. R. Zamir and M. Shah. Accurate image localization based on google maps street view. In *Computer Vision–ECCV 2010*, pages 255–268. Springer, 2010. 3

[40] B. Zhou, A. Lapedriza, J. Xiao, A. Torralba, and A. Oliva. Learning deep features for scene recognition using places database. In *Advances in Neural Information Processing Systems*, pages 487–495, 2014. 2

# Supplemental Material

## 1. Implementation

Here we will provide some more detail about our implementation. We use Caffe framework [13]. We use learning rate 0.01 and reduce it several time during the training, to 0.00001 (when the loss seems to stop improving). Mini-batch size is 32, momentum is 0.9 and weight decay factor is 0.0005.

We use VGG trained on ImageNet [28] as initialization and train a network with the 1060 ways classification for 500k iterations. Then we use this network as initialization for training every other networks (usually just another 100k-200k iterations), we found that this speed up the experiment quite a lot since training every model from scratch or ImageNet initialization take much more time. As shown in Table 3, the pretrained ImageNet model ([I]) can be also be used for retrieval, but not as effective as a model trained for geolocalization task ([L]).

When training with multiple losses, the overall loss will be the weighted sum of all the losses. For [M] model, we use the same weight (1) for all 6 losses.

## 2. Feature visualization

We show a t-SNE visualization in Figure 9. The feature learnt from GPS supervision seems to be very high level; there's many regions in this visualization with consistent theme such as: sport scene images, people images, beach and sunset images, animal images, landmark type of architecture images, etc. There's a large variety in image appearance within a region.

In Figure 10 we look at some dimensions in the output feature space and show the images whose has a high corresponding feature value. Few activation outputs do correspond to some particularly popular landmarks/architecture; while many correspond to certain type of scene or visual features. Some seems to respond to more than one visual features and some might roughly represent higher level location-based semantics. For example row 5 shows pictures of Disney-like castle and Disney's Mickey mouse even-though they are not visually similar.

We show some more nearest neighbors example result in Figure 11.

Table 3. Performance on Im2GPS3k test set.

| Method | Model | Stre. | City | Reg. | Cou. | Cont. |
|---|---|---|---|---|---|---|
| NN | [I] | 7.4 | 17.0 | 19.6 | 26.8 | 41.9 |
| | [L] | 7.5 | 18.9 | 23.5 | 32.6 | 49.5 |
| kNN,$\sigma$=1 | [I] | 7.5 | 18.3 | 22.5 | 30.2 | 45.8 |
| | [L] | 7.8 | 20.9 | 27.1 | 36.8 | 53.8 |
| kNN,$\sigma$=4 | [I] | 7.0 | 16.8 | 22.1 | 31.9 | 48.7 |
| | [L] | 7.2 | 19.4 | 26.9 | 38.9 | 55.9 |
| kNN,$\sigma$=16 | [I] | 4.4 | 10.6 | 15.4 | 32.2 | 51.2 |
| | [L] | 5.3 | 13.8 | 21.2 | 39.9 | 58.9 |

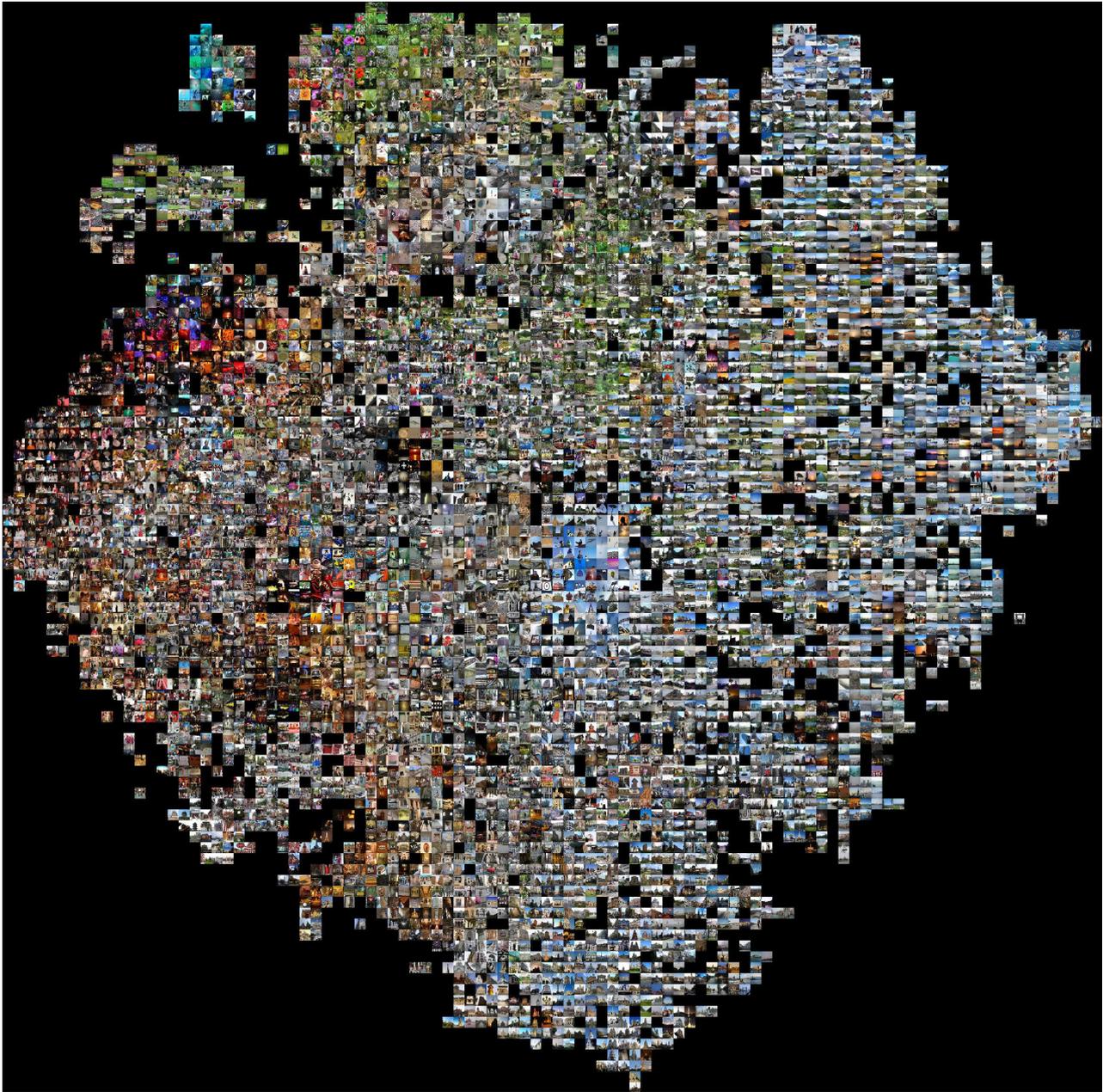

Figure 9. t-SNE visualization

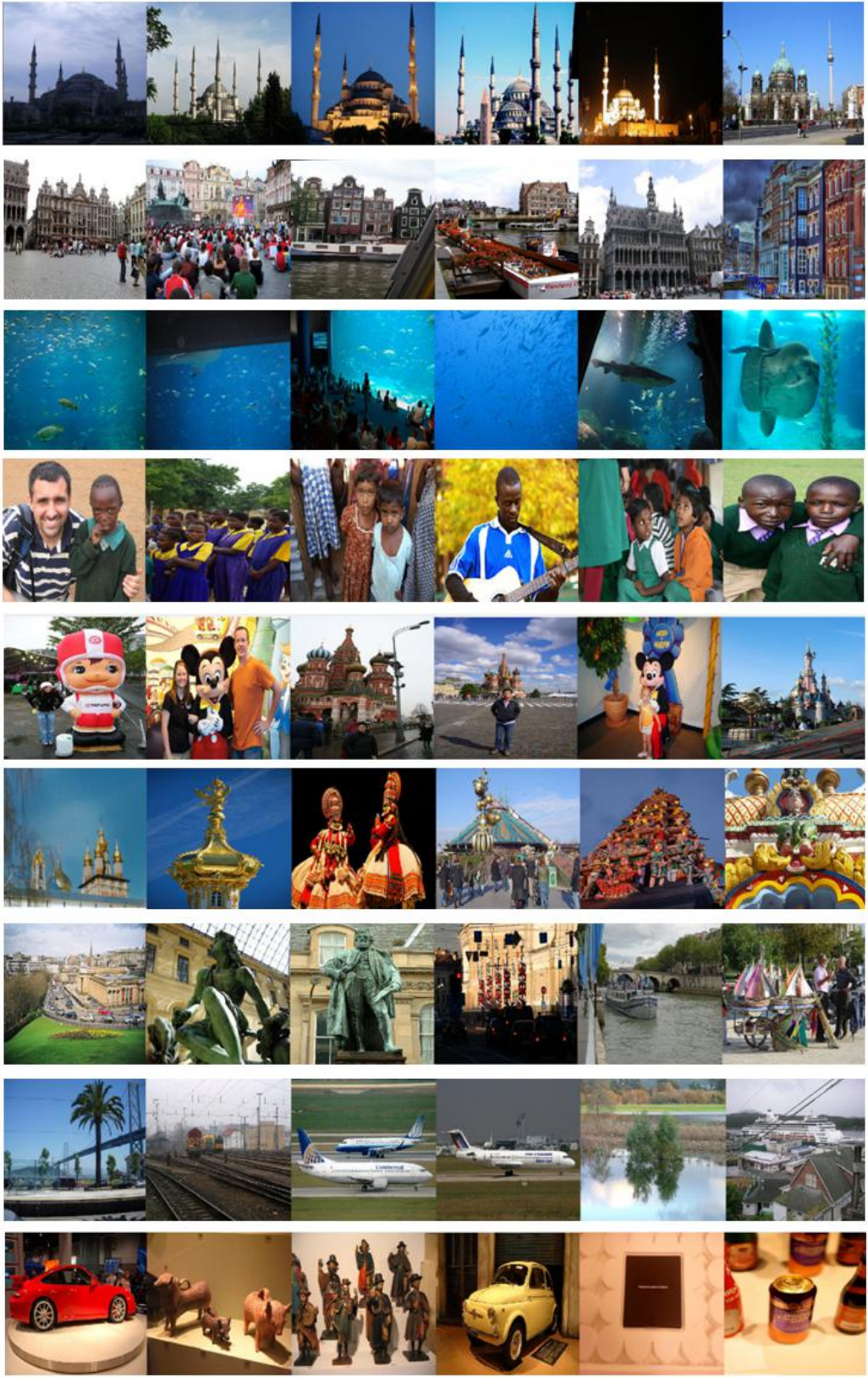

Figure 10. Each row shows a set of images whose feature has a high value at a particular activiation unit (last layer).

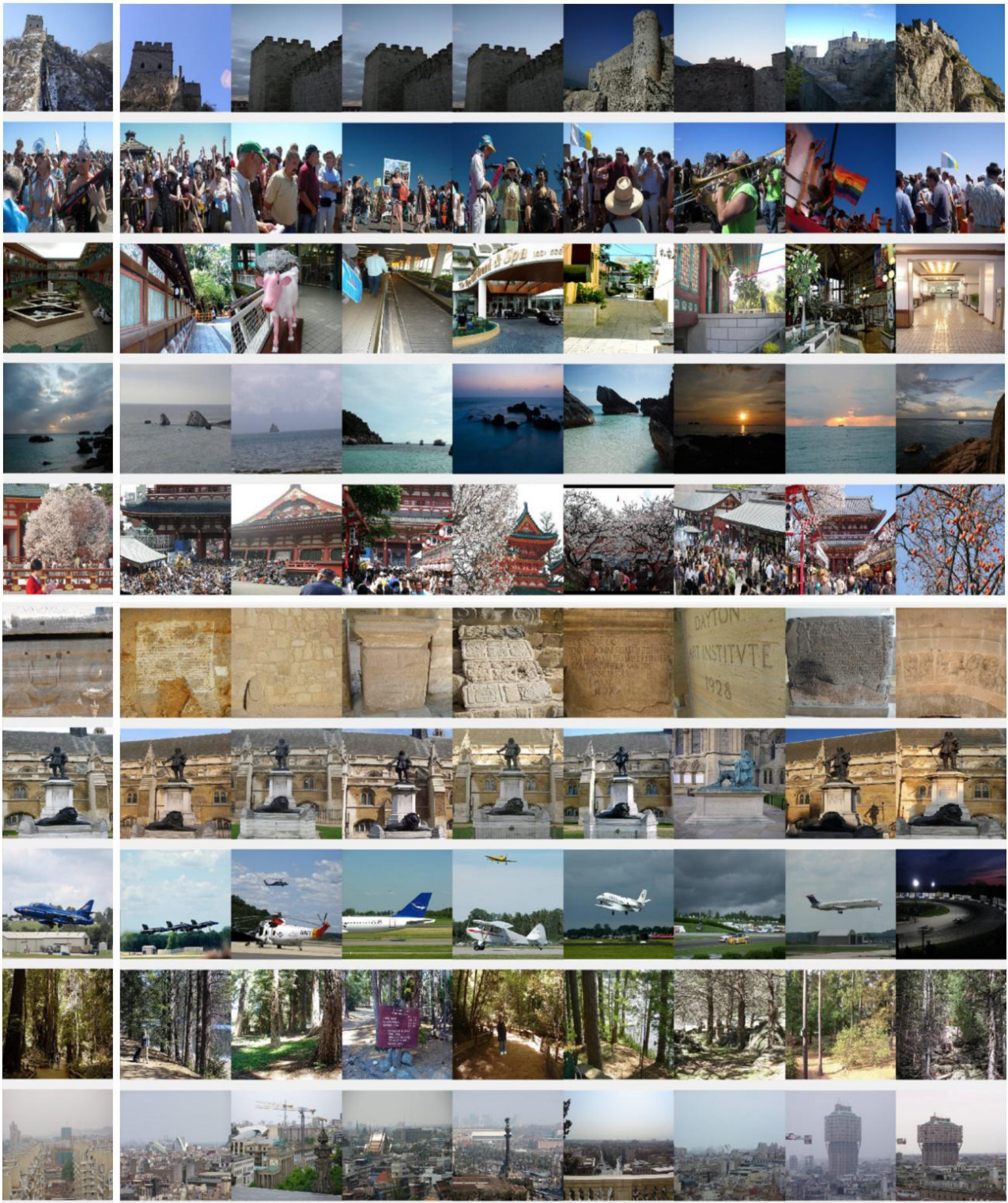

Figure 11. Some qualitative near neighbors result: the images on the left column are query, the other on the same row are its NNs.